\documentclass[a4paper,11pt]{article}
\usepackage{acl2013}
\usepackage[T1]{fontenc}
\usepackage[utf8]{inputenc}

\usepackage{float}
\usepackage{graphicx}
\usepackage{xargs}
\usepackage{algorithm}
\usepackage{algpseudocode}
\usepackage{amsmath,amssymb,amsthm}
\usepackage{stmaryrd}
\usepackage[english]{babel}

\usepackage{rotating}

\usepackage{pgf}
\usepackage{tikz}
\usetikzlibrary{automata,positioning,shapes,arrows}

\makeatletter
\def\th@plain{%
  \thm@notefont{}
  \itshape 
}
\def\th@definition{%
  \thm@notefont{}
  \normalfont 
}
\makeatother

\theoremstyle{theorem}

\newtheorem{definition}{Definition}[section]
\newtheorem{example}{Example}[section]

\newcommand{\ssl}{{\mathsf{SL}}}
\newcommand{\pdlsl}{$\mathrm{PDL}_\ssl$ }

\newcommandx{\atom}[3][2=,3=]{\ensuremath{\mathbf{#1}_\mathsf{#2}^\mathsf{#3}}}

\newcommandx{\mcal}[2]{\ensuremath{\mathcal{#1}_{#2}}}
\newcommandx{\m}[2][1=,2=]{\ensuremath{\mathcal{M}_{#1}^{#2}}}
\newcommandx{\h}[2][1=,2=]{\ensuremath{\mathcal{H}_{#1}^{#2}}}

\newcommandx{\map}[2][1=,2=\cdot]{
    \ensuremath{
        \llbracket #2 \rrbracket_{#1}
    }
}

\title{Sign Language Lexical Recognition With Propositional Dynamic Logic }
\author{Arturo Curiel \thanks{~Supported by CONACYT (Mexico) scholarship program.} \\
  Universit\'e Paul Sabatier\\
  118 route de Narbonne, IRIT,\\
  31062, Toulouse, France \\
  {\tt curiel@irit.fr} \\\And
  Christophe Collet \\
  Universit\'e Paul Sabatier\\
  118 route de Narbonne, IRIT,\\
  31062, Toulouse, France \\
  {\tt collet@irit.fr} \\}
  
\date{}

\begin{document}
    \maketitle
    \begin{abstract}
        This paper explores the use of Propositional Dynamic Logic (PDL) as a suitable formal framework for describing Sign Language (SL), the language of deaf people, in the context of natural language processing. SLs are visual, complete, standalone languages which are just as expressive as oral languages. Signs in SL usually correspond to sequences of highly specific body postures interleaved with movements, which make reference to real world objects, characters or situations. Here we propose a formal representation of SL signs, that will help us with the analysis of automatically-collected hand tracking data from French Sign Language (FSL) video corpora. We further show how such a representation could help us with the design of computer aided SL verification tools, which in turn would bring us closer to the development of an automatic recognition system for these languages.
    \end{abstract}
    
    \section{Introduction}\label{introduction}
        
        Sign languages (SL), the vernaculars of deaf people, are complete, rich, standalone communication systems which have evolved in parallel with oral languages \cite{valli_linguistics_2000}. However, in contrast to the last ones, research in automatic SL processing has not yet managed to build a complete, formal definition oriented to their automatic recognition \cite{cuxac_problematique_2007}. In SL, both hands and non-manual features (NMF), {\em e.g.} facial muscles, can convey information with their placements, configurations and movements. These particular conditions can difficult the construction of a formal description with common natural language processing (NLP) methods, since the existing modeling techniques are mostly designed to work with one-channel sound productions inherent to oral languages, rather than with the multi-channel partially-synchronized information induced by SLs.

        Our research strives to address the formalization problem by introducing a logical language that lets us represent SL from the lowest level, so as to render the recognition task more approachable. For this, we use an instance of a formal logic, specifically Propositional Dynamic Logic (PDL), as a possible description language for SL signs.

        For the rest of this section, we will present a brief introduction to current research efforts in the area. Section~\ref{signlanguagelogic} presents a general description of our formalism, while section~\ref{proofofconcept} shows how our work can be used when confronted with real world data. Finally, section~\ref{conclusions_fw} present our final observations and future work. 
        
        Images for the examples where taken from \cite{dictasign} corpus.
        
        \subsection{Current Sign Language Research}

            Extensive efforts have been made to achieve efficient automatic capture and representation of the subtle nuances commonly present in sign language discourse~\cite{ong_automatic_2005}. Research ranges from the development of hand and body trackers~\cite{dreuw_enhancing_2009,gianni_robust_2009}, to the design of high level SL representation models~\cite{lejeune_analyse_2004,lenseigne_using_2006}. Linguistic research in the area has focused on the characterization of corporal expressions into meaningful transcriptions~\cite{dreuw_signspeak_2010,stokoe_sign_2005} or common patterns across SL~\cite{aronoff_paradox_2005,meir_re-thinking_2006,wittmann_classification_1991}, so as to gain understanding of the underlying mechanisms of SL communication.

            Works like~\cite{losson_sign_1998} deal with the creation of a lexical description oriented to computer-based sign animation. Report \cite{filhol_zebedee:_2009} describes a lexical specification to address the same problem. Both propose a thoroughly geometrical parametric encoding of signs, thus leaving behind meaningful information necessary for recognition and introducing data beyond the scope of recognition. This complicates the reutilization of their formal descriptions. Besides, they don't take in account the presence of partial information. Treating partiality is important for us, since it is often the case with automatic tools that incomplete or unrecognizable information arises. Finally, little to no work has been directed towards the unification of raw collected data from SL corpora with higher level descriptions~\cite{dalle_high_2006}.

    \section{Propositional Dynamic Logic for SL}\label{signlanguagelogic}
 
        {\em Propositional Dynamic Logic} (PDL) is a multi-modal logic, first defined by \cite{fischer_propositional_1979}. It provides a language for describing programs, their correctness and termination, by allowing them to be modal operators. We work with our own variant of this logic, the {\em Propositional Dynamic Logic for Sign Language} ($\mathrm{PDL}_{\mathsf{SL}}$), which is just an instantiation of PDL where we take signers' movements as programs.

        Our sign formalization is based on the approach of \cite{liddell_american_1989} and \cite{filhol_modedescriptif_2008}. They describe signs as sequences of immutable {\em key postures} and movement {\em transitions}. 
          
        In general, each key posture will be characterized by the concurrent parametric state of each {\em body articulator} over a time-interval. For us, a body articulator is any relevant body part involved in signing. The parameters taken in account can vary from articulator to articulator, but most of the time they comprise their configurations, orientations and their placement within one or more {\em places of articulation}. Transitions will correspond to the movements executed between fixed postures. 
    
        \subsection{Syntax}
    
            We need to define some primitive sets that will limit the domain of our logical language.
    
\begin{figure}
                \begin{center}
                \includegraphics[scale=.45]{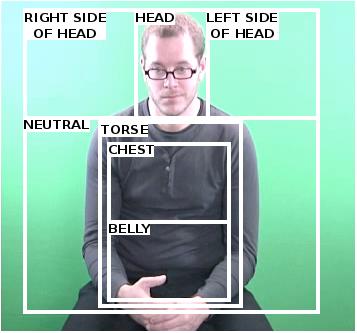}
                \end{center}
                \caption{Possible places of articulation in $\mathcal{B}_\mathsf{SL}$. \label{fig:neutralspace}}
                \end{figure}

            \begin{definition}[Sign Language primitives]
    
                Let $\mathcal{B}_\mathsf{SL} = \{\mathbb{D, W, R, L}\}$ be the set of {\em relevant body articulators for SL}, where $\mathbb{D}$, $\mathbb{W}$, $\mathbb{R}$ and $\mathbb{L}$ represent the dominant, weak, right and left hands, respectively. Both $\mathbb{D}$ and $\mathbb{W}$ can be aliases for the right or left hands, but they change depending on whether the signer is right-handed or left-handed, or even depending on the context.

                Let $\Psi$ be the two-dimensional projection of a human body skeleton, seen by the front. We define the set of {\em places of articulation for SL} as $\Lambda_\mathsf{SL} = \{ \mathtt{HEAD}, \mathtt{CHEST}, \mathtt{NEUTRAL}, \ldots \}$, such that for each $\lambda\in\Lambda_\mathsf{SL}$, $\lambda$ is a sub-plane of $\Psi$, as shown graphically in figure~\ref{fig:neutralspace}.
                
                Let $\mathbb{C}_\mathsf{SL}$ be the set of possible morphological configurations for a hand.
    
                Let $\Delta = \{\uparrow,\nearrow, \rightarrow,\searrow,\downarrow,\swarrow,\leftarrow,\nwarrow\}$ be the set of {\em relative directions} from the signer's point of view, where each arrow represents one of eight possible two-dimensional direction vectors that share the same origin. For vector $\delta \in \Delta$, we define vector $\overleftarrow{\delta}$ as the same as $\delta$ but with the inverted abscissa axis, such that $\overleftarrow{\delta} \in \Delta$. Let vector $\widehat{\delta}$ indicate movement with respect to the dominant or weak hand in the following manner:
                  \[\widehat{\delta} = \left\{ 
                      \begin{array}{rl}
                          \delta &\mbox{ if $~\mathbb{D} \equiv \mathbb{R}~$ or $~\mathbb{W} \equiv \mathbb{L}$} \\
                          \overleftarrow{\delta} &\mbox{ if $~\mathbb{D} \equiv \mathbb{L}~$ or $~\mathbb{W} \equiv \mathbb{R}$}
                      \end{array}
                      \right.
                  \]          
                
                Finally, let $\overrightarrow{v_1}$ and $\overrightarrow{v_2}$ be any two vectors with the same origin. We denote the rotation angle between the two as $\theta(\overrightarrow{v_1}, \overrightarrow{v_2})$.
                
            \end{definition}
    
            Now we define the set of atomic propositions that we will use to characterize fixed states, and a set of atomic actions to describe movements.
            
            \begin{definition}[Atomic Propositions for SL Body Articulators $\Phi_{\mathsf{SL}}$]\label{atomicdefinitions}
            The set of {\em atomic propositions for SL articulators} ($\Phi_{\mathsf{SL}}$) is defined as:
        
            \[\Phi_{\mathsf{SL}}=\{{\beta_1}^{\delta}_{\beta_2}, 
                                \Xi^{\beta_1}_\lambda, 
                                \mathcal{T}^{\beta_1}_{\beta_2}, 
                                \mathcal{F}^{\beta_1}_c, 
                                \angle^\delta_{\beta_1}
                              \}\]
                              
            where ${\beta_1}, {\beta_2} \in \mathcal{B}_\mathsf{SL}$, $\delta \in \Delta$, $\lambda \in \Lambda_\mathsf{SL}$ and $c \in \mathbb{C}_\mathsf{SL}$.

            Intuitively, ${\beta_1}^{\delta}_{\beta_2}$ indicates that articulator ${\beta_1}$ is placed in relative direction $\delta$ with respect to articulator ${\beta_2}$. Let the current place of articulation of ${\beta_2}$ be the origin point of ${\beta_2}$'s Cartesian system ($\mathcal{C}_{\beta_2}$). Let vector $\overrightarrow{{\beta_1}}$ describe the current place of articulation of ${\beta_1}$ in $\mathcal{C}_{\beta_2}$. Proposition ${\beta_1}^{\delta}_{\beta_2}$ holds when $\forall \overrightarrow{v} \in \Delta$, $\theta(\overrightarrow{{\beta_1}}, \delta) \leq \theta(\overrightarrow{{\beta_1}}, \overrightarrow{v})$. 
    
            $\Xi^{\beta_1}_\lambda$ asserts that articulator ${\beta_1}$ is located in $\lambda$.
            
            $\mathcal{T}^{\beta_1}_{\beta_2}$ is active whenever articulator ${\beta_1}$ physically touches articulator ${\beta_2}$.
    
            $\mathcal{F}^{\beta_1}_c$ indicates that $c$ is the morphological configuration of articulator ${\beta_1}$. 
        
            Finally, $\angle^\delta_{\beta_1}$ means that an articulator ${\beta_1}$ is {\em oriented} towards direction $\delta \in \Delta$. For hands, $\angle^\delta_{\beta_1}$ will hold whenever the vector perpendicular to the plane of the palm has the smallest rotation angle with respect to $\delta$.
            \end{definition}

            \begin{definition}[Atomic Actions for SL Body Articulators $\Pi_\mathsf{SL}$]
               The {\em atomic actions for SL articulators} (~$\Pi_\mathsf{SL}$) are given by the following set:
             
             \[\Pi_\mathsf{SL}=\{\delta_{\beta_1}, \leftrightsquigarrow_{\beta_1}\}\]
        where $\delta \in \Delta$  and ${\beta_1} \in \mathcal{B}_\mathsf{SL}$. 
            
               Let ${\beta_1}$'s position before movement be the origin of ${\beta_1}$'s Cartesian system ($\mathcal{C}_{\beta_1}$) and $\overrightarrow{{\beta_1}}$ be the position vector of ${\beta_1}$ in $\mathcal{C}_{\beta_1}$ after moving. Action $\delta_{\beta_1}$ indicates that ${\beta_1}$ moves in relative direction $\delta$ in $\mathcal{C}_{\beta_1}$ if ~$\forall \overrightarrow{v} \in \Delta$, $\theta(\overrightarrow{{\beta_1}},\delta) \leq \theta(\overrightarrow{{\beta_1}},\overrightarrow{v})$.
        
            Action $\leftrightsquigarrow_{\beta_1}$ occurs when articulator ${\beta_1}$ moves rapidly and continuously ({\em thrills}) without changing it's current place of articulation.
            \end{definition}

            \begin{definition}[Action Language for SL Body Articulators $\mathcal{A}_\mathsf{SL}$]\label{actionlanguage}
            The {\em action language for body articulators} ($\mathcal{A}_\mathsf{SL}$) is given by the following rule:
             
             \[\alpha::=\pi~|~\alpha\cap\alpha~|~\alpha\cup\alpha~|~\alpha;\alpha~|~\alpha^*\]
        where $\pi \in \Pi_\mathsf{SL}$.
        
             Intuitively, $\alpha\cap\alpha$ indicates the concurrent execution of two actions, while $\alpha\cup\alpha$ means that at least one of two actions will be non-deterministically executed. Action $\alpha;\alpha$ describes the sequential execution of two actions. Finally, action $\alpha^*$ indicates the reflexive transitive closure of $\alpha$.
            \end{definition}
            
            \begin{definition}[Language \pdlsl]\label{languagedefinition}
            The formulae $\varphi$ of \pdlsl are given by the following rule:
        
            \[\varphi ::= \top~|~p~|~\neg\varphi~|~\varphi \wedge \varphi~|~[\alpha]\varphi\]
    
            where $p \in \Phi_{\mathsf{SL}}$, $\alpha \in \mathcal{A}_\mathsf{SL}$.
            \end{definition}
    
        \subsection{Semantics}
    
            \pdlsl~formulas are interpreted over labeled transition systems (LTS), in the spirit of the possible worlds model introduced by ~\cite{hintikka_knowledge_1962}. Models correspond to connected graphs representing key postures and transitions: states are determined by the values of their propositions, while edges represent sets of executed movements. Here we present only a small extract of the logic semantics.
    
            \begin{definition}[Sign Language Utterance Model $\mathcal{U}_\mathsf{SL}$]\label{kripkedefinition}
            A {\em sign language utterance model} ($\mathcal{U}_\mathsf{SL}$), is a tuple $\mathcal{U}_\mathsf{SL}=(S, R, \llbracket \cdot \rrbracket_{\Pi_{\mathsf{SL}}}, \llbracket \cdot \rrbracket_{\Phi_{\mathsf{SL}}})$ where:
            
                \begin{itemize}
                    \item $S$ is a non-empty set of states
                    \item $R$ is a transition relation $R \subseteq S \times S$ where, $\forall s\in~S, \exists s'\in S \text{ such that } (s,s')\in R$.
                    \item $\llbracket \cdot \rrbracket_{\Pi_{\mathsf{SL}}}:~\Pi_{\mathsf{SL}} \rightarrow R$, denotes the function mapping actions to the set of binary relations.
                    \item $\llbracket \cdot \rrbracket_{\Phi_{\mathsf{SL}}}:~S\rightarrow 2^{\Phi_{\mathsf{SL}}}$, maps each state to a set of atomic propositions.
                \end{itemize}            
            \end{definition}
            
            We also need to define a structure over sequences of states to model internal dependencies between them, nevertheless we decided to omit the rest of our semantics, alongside satisfaction conditions, for the sake of readability.

    \section{Use Case: Semi-Automatic Sign Recognition}\label{proofofconcept}

\begin{figure}[htb]
                \begin{tikzpicture}[auto, node distance=1.6cm, >=latex', inner sep=0pt]

                    \tikzstyle{tool} = [fill=blue!20, rectangle, minimum width=1.4cm, minimum height=1.4cm]
                    \tikzstyle{resource} = [fill=red!20, circle, minimum width=1.2cm]
                    \tikzstyle{output} = [coordinate, node distance=1.6cm]
                    \tikzstyle{tag} = [text width=4.6em, font=\scriptsize\sffamily, align=center]

                    \node[resource] (corpus) {};
                    \node[tool, right of=corpus](tracking){};
                    \node[resource, below of=tracking, minimum width=1.4cm](segmentation){};
                    \node[tool, right of=segmentation](modeliser){};
                    \node[tool, right of=modeliser](checker){};
                    \node[resource, above of=modeliser](model){};
                    \node[resource, above of=checker](formules){};
                    \node[resource, right of=formules](rules){};
                    \node[resource, below of=rules](lexicaldata){};

                    \node[tag, right=15pt of corpus, anchor=east](corpustag) {Corpus};
                    \node[tag, right=12pt of tracking, anchor=east](trackingtag){Tracking and Segmentation Module};    
                    \node[tag, right=12pt of segmentation, anchor=east](segmentationtag) {Key postures~\& transitions};
                    \node[tag, right=12pt of modeliser, anchor=east](modelisertag){\pdlsl \\ Model Extraction Module};    
                    \node[tag, right=12pt of checker, anchor=east](checkertag) {\pdlsl \\ Verification Module};
                    \node[tag, right=15pt of model, anchor=east](modeltag){\pdlsl \\Graph};
                    \node[tag, right=15pt of formules, anchor=east](formulestag){Sign Formulæ};
                    \node[tag, right=15pt of rules, anchor=east](rulestag){User \\ Input};
                    \node[tag, right=15pt of lexicaldata, anchor=east](lexicaldatatag){Sign Proposals};
                       
                    \path[->] (corpus) edge (tracking)
                              (tracking) edge (segmentation)
                              (segmentation) edge (modeliser)
                                 (modeliser) edge (model)
                              (model) edge (checker)
                              (formules) edge (checker)
                                (rules) edge (checker)
                              (checker) edge (lexicaldata);
                \end{tikzpicture}
                \caption{Information flow in a semi-automatic SL lexical recognition system.}
                \label{architecture}
            \end{figure}
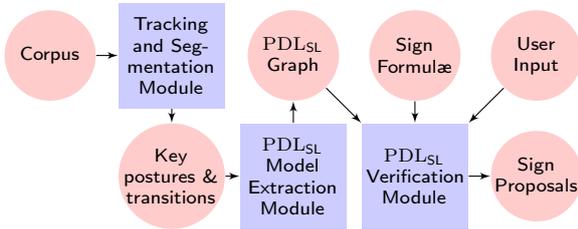

        We now present an example of how we can use our formalism in a semi-automatic sign recognition system. Figure \ref{architecture} shows a simple module diagram exemplifying information flow in the system's architecture. We proceed to briefly describe each of our modules and how they work together.

            \subsection{Tracking and Segmentation Module}
            
               The process starts by capturing relevant information from video corpora. We use an existing head and hand tracker expressly developed for SL research~\cite{gonzalez_robust_2011}. This tool analyses individual video instances, and returns the frame-by-frame positions of the tracked articulators. By using this information, the module can immediately calculate speeds and directions on the fly for each hand. 
    
               The module further employs the method proposed by the authors in~\cite{gonzalez_sign_2012} to achieve sub-lexical segmentation from the previously calculated data. Like them, we use the relative velocity between hands to identify when hands either move at the same time, independently or don't move at all. With these, we can produce a set of possible key postures and transitions that will serve as input to the modeling module.

           \subsection{Model Extraction Module}

               This module calculates a propositional state for each static posture, where atomic \pdlsl formulas codify the information tracked in the previous part. Detected movements are interpreted as \pdlsl actions between states.

               \begin{figure}[htb]
                   \centering
                   \begin{minipage}[b]{0.98\columnwidth}
                       \centering
                       \includegraphics[width=0.23\columnwidth]{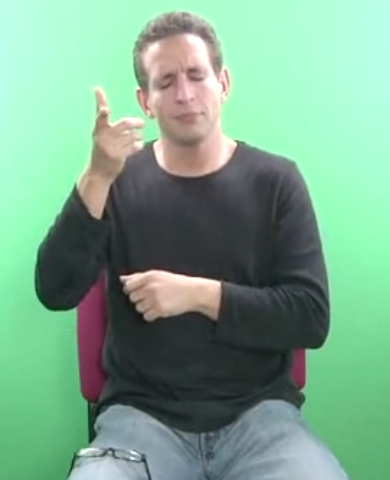}
                       \includegraphics[width=0.23\columnwidth]{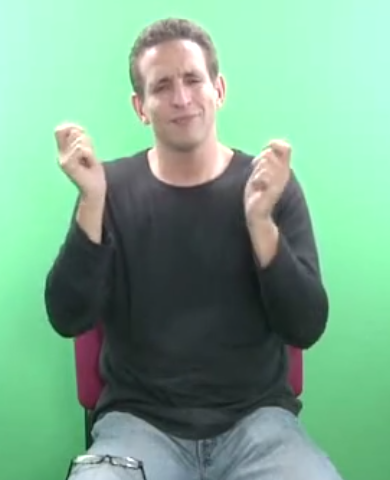}
                       \includegraphics[width=0.23\columnwidth]{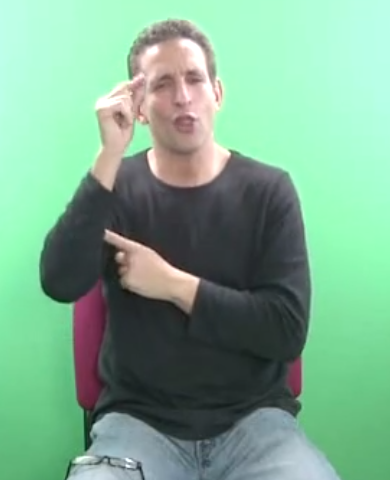}
                       \includegraphics[width=0.23\columnwidth]{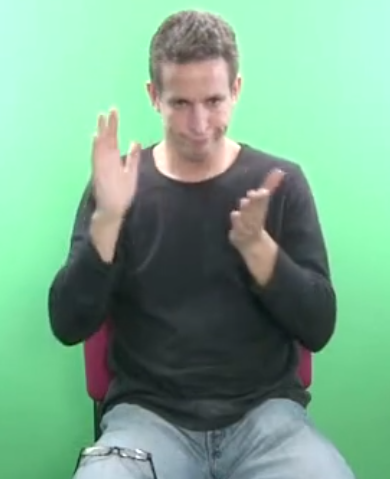}
                   \end{minipage} \\
                   \begin{minipage}[b]{.9\columnwidth}
                   \centering
                       \begin{tikzpicture}[->,>=stealth',shorten >=1pt,auto, font=\scriptsize]
                          \tikzstyle{every state}=[circle, fill=blue!65, draw=none, text=white,
                          text width=.2cm, inner sep=0pt, minimum size=10pt, node distance=1.75cm]
                          \tikzstyle{invisible}=[fill=none,draw=none,text=black, node distance=.4cm]
                          
                          \node[state] (s0) {$\vdots$};
                          \node[invisible] (s0tag1) [below=0.05cm of s0] {$\mathbb{R}^\nearrow_{\mathbb{L}}$};
                          \node[invisible] (s0tag2) [below of=s0tag1] {$\Xi^\mathbb{L}_{\mathtt{TORSE}}$};
                          \node[invisible] (s0tag3) [below of=s0tag2] {$\Xi^\mathbb{R}_{\mathtt{R\_SIDEOFBODY}}$};
                          \node[invisible] (s0tag4) [below of=s0tag3] {$\neg\mathcal{F}^{\mathbb{R}}_{\mathtt{L\_CONFIG}}$};
                          \node[invisible] (s0tag5) [below of=s0tag4] {$\neg\mathcal{F}^{\mathbb{L}}_{\mathtt{FIST\_CONFIG}}$};
                          \node[invisible] (s0tag6) [below of=s0tag5] {$\neg\mathcal{T}^{\mathbb{R}}_{\mathbb{L}}$};
                          \node[invisible] (s0tag7) [below of=s0tag6] {$\vdots$};
                          \node[state] (s1) [right of=s0] {$\vdots$};
                          \node[invisible] (s1tag1) [below=0.05cm of s1] {$\mathbb{R}^\leftarrow_{\mathbb{L}}$};
                          \node[invisible] (s1tag2) [below of=s1tag1] {$\Xi^\mathbb{L}_{\mathtt{L\_SIDEOFBODY}}$};
                          \node[invisible] (s1tag3) [below of=s1tag2] {$\Xi^\mathbb{R}_{\mathtt{R\_SIDEOFBODY}}$};
                          \node[invisible] (s1tag4) [below of=s1tag3] {$\mathcal{F}^{\mathbb{R}}_{\mathtt{KEY\_CONFIG}}$};
                          \node[invisible] (s1tag5) [below of=s1tag4] {$\mathcal{F}^{\mathbb{L}}_{\mathtt{KEY\_CONFIG}}$};
                          \node[invisible] (s1tag6) [below of=s1tag5] {$\neg\mathcal{T}^{\mathbb{R}}_{\mathbb{L}}$};
                          \node[invisible] (s1tag7) [below of=s1tag6] {$\vdots$};
                          \path[->] (s0) edge node {$\nearrow_{\mathbb{L}}$} (s1);
                          \path[->] (s1) edge[loop above, distance=.5cm] node {$\leftrightsquigarrow_\mathbb{D} \cap \leftrightsquigarrow_\mathbb{G}$} (s1);
                          \node[state] (s2) [right of=s1] {$\vdots$};
                          \node[invisible] (s2tag1) [below=0.05cm of s2] {$\mathbb{R}^\leftarrow_{\mathbb{L}}$};
                          \node[invisible] (s2tag2) [below of=s2tag1] {$\Xi^\mathbb{L}_{\mathtt{CENTEROFBODY}}$};
                          \node[invisible] (s2tag3) [below of=s2tag2] {$\Xi^\mathbb{R}_{\mathtt{R\_SIDEOFHEAD}}$};
                          \node[invisible] (s2tag4) [below of=s2tag3] {$\mathcal{F}^{\mathbb{R}}_{\mathtt{BEAK\_CONFIG}}$};
                          \node[invisible] (s2tag5) [below of=s2tag4] {$\mathcal{F}^{\mathbb{L}}_{\mathtt{INDEX\_CONFIG}}$};
                          \node[invisible] (s2tag6) [below of=s2tag5] {$\neg\mathcal{T}^{\mathbb{R}}_{\mathbb{L}}$};                          
                          \node[invisible] (s2tag7) [below of=s2tag6] {$\vdots$};
                          \path[->] (s1) edge node {$\swarrow_{\mathbb{L}}$} (s2);
                          \node[state] (s3) [right of=s2] {$\vdots$};
                          \node[invisible] (s3tag1) [below=0.05cm of s3] {$\mathbb{R}^\leftarrow_{\mathbb{L}}$};
                          \node[invisible] (s3tag2) [below of=s3tag1] {$\Xi^\mathbb{L}_{\mathtt{L\_SIDEOFBODY}}$};
                          \node[invisible] (s3tag3) [below of=s3tag2] {$\Xi^\mathbb{R}_{\mathtt{R\_SIDEOFBODY}}$};
                          \node[invisible] (s3tag4) [below of=s3tag3] {$\mathcal{F}^{\mathbb{R}}_{\mathtt{OPENPALM\_CONFIG}}$};
                          \node[invisible] (s3tag5) [below of=s3tag4] {$\mathcal{F}^{\mathbb{L}}_{\mathtt{OPENPALM\_CONFIG}}$};
                          \node[invisible] (s3tag6) [below of=s3tag5] {$\neg\mathcal{T}^{\mathbb{R}}_{\mathbb{L}}$};                          
                          \node[invisible] (s3tag7) [below of=s3tag6] {$\vdots$};
                          \path[->] (s2) edge node {$\nearrow_{\mathbb{L}}$} (s3);
                        \end{tikzpicture}
                   \end{minipage}
                   \caption{Example of modeling over four automatically identified frames as possible key postures.}
                   \label{fig:modeling}
               \end{figure}
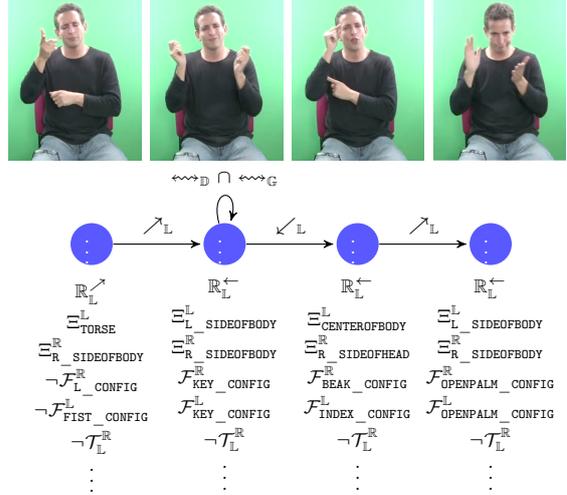

               Figure ~\ref{fig:modeling} shows an example of the process. Here, each key posture is codified into propositions acknowledging the hand positions with respect to each other ($\mathbb{R}^\leftarrow_{\mathbb{L}}$), their place of articulation ({\em e.g.} ``left hand floats over the torse'' with $\Xi^\mathbb{L}_{\mathtt{TORSE}}$), their configuration ({\em e.g.} ``right hand is open'' with $\mathcal{F}^{\mathbb{R}}_{\mathtt{OPENPALM\_CONFIG}}$) and their movements ({\em e.g.} ``left hand moves to the up-left direction'' with $\nearrow_{\mathbb{L}}$).
               
               This module also checks that the generated graph is correct: it will discard simple tracking errors to ensure that the resulting LTS will remain consistent.
               
%
%

           \subsection{Verification Module}

               First of all, the verification module has to be loaded with a database of sign descriptions encoded as \pdlsl formulas. These will characterize the specific sequence of key postures that morphologically describe a sign. For example, let's take the case for sign ``route'' in FSL, shown in figure~\ref{fig:routelsf}, with the following \pdlsl formulation,
   
               \begin{example}[$\mathtt{ROUTE}_\mathsf{FSL}$ formula]
                   \begin{equation}
                       \begin{split}
                           (\Xi^\mathbb{R}_\mathtt{FACE} \wedge
                           \Xi^\mathbb{L}_\mathtt{FACE} \wedge
                           \mathbb{L}^\rightarrow_\mathbb{R} \wedge
                           \mathcal{F}^\mathbb{R}_\mathtt{CLAMP} \wedge
                           \mathcal{F}^\mathbb{L}_\mathtt{CLAMP} \wedge
                           \mathcal{T}^\mathbb{R}_\mathbb{L})
                           \rightarrow \\
                           [\leftarrow_\mathbb{R} \cap
                                 \rightarrow_\mathbb{L}](
                                 \mathbb{L}^\rightarrow_\mathbb{R} \wedge
                                 \mathcal{F}^\mathbb{R}_\mathtt{CLAMP} \wedge
                                 \mathcal{F}^\mathbb{L}_\mathtt{CLAMP} \wedge
                                 \neg\mathcal{T}^\mathbb{R}_\mathbb{L})
                      \end{split}
                      \label{eq:route}
                  \end{equation}
                 
                  \begin{figure}[htb]
                      \begin{center}
                      \includegraphics[scale=0.26]{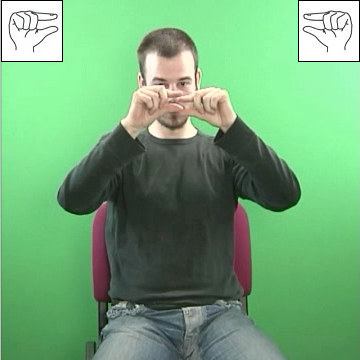}
                      \includegraphics[scale=0.26]{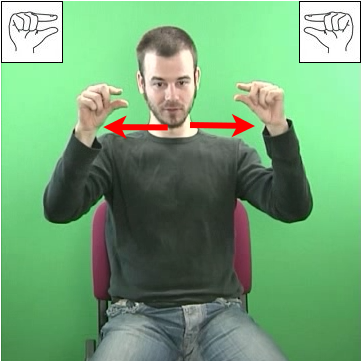}
                      \end{center}            
                      \caption{$\mathtt{ROUTE}_\mathsf{FSL}$ production. \label{fig:routelsf}}
                  \end{figure}
               \end{example}
               
            Formula (\ref{eq:route}) describes $\mathtt{ROUTE}_\mathsf{FSL}$ as a sign with two key postures, connected by a two-hand simultaneous movement (represented with operator $\cap$). It also indicates the position of each hand, their orientation, whether they touch and their respective configurations (in this example, both hold the same \texttt{CLAMP} configuration).

            The module can then verify whether a sign formula in the lexical database holds in any sub-sequence of states of the graph generated in the previous step. Algorithm \ref{alg:verification} sums up the process.

               \begin{algorithm}
                   \caption{\pdlsl Verification Algorithm}
                   \label{alg:verification}
                   \begin{algorithmic}[1]
                       \Require SL model $\mathcal{M}_\ssl$
                       \Require connected graph $\mathcal{G}_\ssl$
                       \Require lexical database $\mathcal{DB}_\ssl$
                       \State Proposals\_For[$state\_qty$]
                       \For{state $s \in \mathcal{G}_\ssl$}
                           \For{sign $\varphi \in \mathcal{DB}_\ssl$ where $s \in \varphi$}
                               \If{$\mathcal{M}_\ssl, s \models \varphi$}
                                   \State Proposals\_For[$s$].append($\varphi$)
                               \EndIf
                           \EndFor
                       \EndFor \\
                       \Return{Proposals\_For}
                   \end{algorithmic}
               \end{algorithm}
               
               For each state, the algorithm returns a set of possible signs. Expert users (or higher level algorithms) can further refine the process by introducing additional information previously missed by the tracker.
 
    \section{Conclusions and Future Work}              
        \label{conclusions_fw}
        
        We have shown how a logical language can be used to model SL signs for semi-automatic recognition, albeit with some restrictions. The traits we have chosen to represent were imposed by the limits of the tracking tools we had to our disposition, most notably working with 2D coordinates. With these in mind, we tried to design something flexible that could be easily adapted by computer scientists and linguists alike. Our primitive sets, were intentionally defined in a very general fashion due to the same reason: all of the perceived directions, articulators and places of articulation can easily change their domains, depending on the SL we are modeling or the technological constraints we have to deal with. Propositions can also be changed, or even induced, by existing written sign representation languages such as Zebedee \cite{filhol_modedescriptif_2008} or HamNoSys \cite{hanke_hamnosysrepresenting_2004}, mainly for the sake of extendability. 
        
        From the application side, we still need to create an extensive sign database codified in \pdlsl and try recognition on other corpora, with different tracking information. For verification and model extraction, further optimizations are expected, including the handling of data inconsistencies and repairing broken queries when verifying the graph.
        
        Regarding our theoretical issues, future work will be centered in improving our language to better comply with SL research. This includes adding new features, like incorporating probability representation to improve recognition. We also expect to finish the definition of our formal semantics, as well as proving correction and complexity of our algorithms.

    \bibliographystyle{acl}
    \bibliography{aclshort}

\begin{thebibliography}{}

\bibitem[\protect\citename{Aronoff \bgroup et al.\egroup
  }2005]{aronoff_paradox_2005}
Mark Aronoff, Irit Meir, and Wendy Sandler.
\newblock 2005.
\newblock The paradox of sign language morphology.
\newblock {\em Language}, 81(2):301.

\bibitem[\protect\citename{Cuxac and Dalle}2007]{cuxac_problematique_2007}
Christian Cuxac and Patrice Dalle.
\newblock 2007.
\newblock {\em Problématique des chercheurs en traitement automatique des
  langues des signes}, volume~48 of {\em Traitement Automatique des Langues}.
\newblock Lavoisier, http://www.editions-hermes.fr/, October.

\bibitem[\protect\citename{Dalle}2006]{dalle_high_2006}
Patrice Dalle.
\newblock 2006.
\newblock High level models for sign language analysis by a vision system.
\newblock In {\em Workshop on the Representation and Processing of Sign
  Language: Lexicographic Matters and Didactic Scenarios ({LREC)}, Italy,
  {ELDA}}, page 17–20.

\bibitem[\protect\citename{DictaSign}2012]{dictasign}
DictaSign.
\newblock 2012.
\newblock http://www.dictasign.eu.

\bibitem[\protect\citename{Dreuw \bgroup et al.\egroup
  }2009]{dreuw_enhancing_2009}
Philippe Dreuw, Daniel Stein, and Hermann Ney.
\newblock 2009.
\newblock Enhancing a sign language translation system with vision-based
  features.
\newblock In Miguel~Sales Dias, Sylvie Gibet, Marcelo~M. Wanderley, and Rafael
  Bastos, editors, {\em Gesture-Based Human-Computer Interaction and
  Simulation}, number 5085 in Lecture Notes in Computer Science, pages
  108--113. Springer Berlin Heidelberg, January.

\bibitem[\protect\citename{Dreuw \bgroup et al.\egroup
  }2010]{dreuw_signspeak_2010}
Philippe Dreuw, Hermann Ney, Gregorio Martinez, Onno Crasborn, Justus Piater,
  Jose~Miguel Moya, and Mark Wheatley.
\newblock 2010.
\newblock The {SignSpeak} project - bridging the gap between signers and
  speakers.
\newblock In Nicoletta Calzolari~{(Conference} Chair), Khalid Choukri, and {\em
  et. al.}, editors, {\em Proceedings of the Seventh International Conference
  on Language Resources and Evaluation {(LREC'10)}}, Valletta, Malta, May.
  European Language Resources Association {(ELRA)}.

\bibitem[\protect\citename{Filhol}2008]{filhol_modedescriptif_2008}
Michael Filhol.
\newblock 2008.
\newblock {\em Modèle descriptif des signes pour un traitement automatique des
  langues des signes}.
\newblock {Ph.D.} thesis, Université Paris-sud {(Paris} 11).

\bibitem[\protect\citename{Filhol}2009]{filhol_zebedee:_2009}
Michael Filhol.
\newblock 2009.
\newblock Zebedee: a lexical description model for sign language synthesis.
\newblock {Internal}, {LIMSI}.

\bibitem[\protect\citename{Fischer and Ladner}1979]{fischer_propositional_1979}
Michael~J. Fischer and Richard~E. Ladner.
\newblock 1979.
\newblock Propositional dynamic logic of regular programs.
\newblock {\em Journal of Computer and System Sciences}, 18(2):194--211, April.

\bibitem[\protect\citename{Gianni and Dalle}2009]{gianni_robust_2009}
Frédéric Gianni and Patrice Dalle.
\newblock 2009.
\newblock Robust tracking for processing of videos of communication’s
  gestures.
\newblock {\em Gesture-Based Human-Computer Interaction and Simulation}, page
  93–101.

\bibitem[\protect\citename{Gonzalez and Collet}2011]{gonzalez_robust_2011}
Matilde Gonzalez and Christophe Collet.
\newblock 2011.
\newblock Robust body parts tracking using particle filter and dynamic
  template.
\newblock In {\em 2011 18th {IEEE} International Conference on Image Processing
  {(ICIP)}}, pages 529 --532, September.

\bibitem[\protect\citename{Gonzalez and Collet}2012]{gonzalez_sign_2012}
Matilde Gonzalez and Christophe Collet.
\newblock 2012.
\newblock Sign segmentation using dynamics and hand configuration for
  semi-automatic annotation of sign language corpora.
\newblock In Eleni Efthimiou, Georgios Kouroupetroglou, and Stavroula-Evita
  Fotinea, editors, {\em Gesture and Sign Language in Human-Computer
  Interaction and Embodied Communication}, number 7206 in Lecture Notes in
  Computer Science, pages 204--215. Springer Berlin Heidelberg, January.

\bibitem[\protect\citename{Hanke}2004]{hanke_hamnosysrepresenting_2004}
Thomas Hanke.
\newblock 2004.
\newblock {HamNoSys—Representing} sign language data in language resources
  and language processing contexts.
\newblock In {\em Proceedings of the Workshop on the Representation and
  Processing of Sign Languages “From SignWriting to Image Processing.
  Information}, Lisbon, Portugal, 30 May.

\bibitem[\protect\citename{Hintikka}1962]{hintikka_knowledge_1962}
Jaakko Hintikka.
\newblock 1962.
\newblock {\em Knowledge and Belief}.
\newblock Ithaca, {N.Y.,Cornell} University Press.

\bibitem[\protect\citename{Lejeune}2004]{lejeune_analyse_2004}
Fanch Lejeune.
\newblock 2004.
\newblock {\em Analyse sémantico-cognitive d'énoncés en Langue des Signes
  Fran{\textbackslash}ccaise pour une génération automatique de séquences
  gestuelles}.
\newblock {Ph.D.} thesis, {PhD} thesis, Orsay University, France.

\bibitem[\protect\citename{Lenseigne and Dalle}2006]{lenseigne_using_2006}
Boris Lenseigne and Patrice Dalle.
\newblock 2006.
\newblock Using signing space as a representation for sign language processing.
\newblock In Sylvie Gibet, Nicolas Courty, and Jean-François Kamp, editors,
  {\em Gesture in Human-Computer Interaction and Simulation}, number 3881 in
  Lecture Notes in Computer Science, pages 25--36. Springer Berlin Heidelberg,
  January.

\bibitem[\protect\citename{Liddell and Johnson}1989]{liddell_american_1989}
S.~K. Liddell and R.~E. Johnson.
\newblock 1989.
\newblock {\em American sign language: The phonological base}.
\newblock Gallaudet University Press, Washington. {DC}.

\bibitem[\protect\citename{Losson and Vannobel}1998]{losson_sign_1998}
Olivier Losson and Jean-Marc Vannobel.
\newblock 1998.
\newblock Sign language formal description and synthesis.
\newblock {\em {INT.JOURNAL} {OF} {VIRTUAL} {REALITY}}, 3:27---34.

\bibitem[\protect\citename{Meir \bgroup et al.\egroup
  }2006]{meir_re-thinking_2006}
Irit Meir, Carol Padden, Mark Aronoff, and Wendy Sandler.
\newblock 2006.
\newblock Re-thinking sign language verb classes: the body as subject.
\newblock In {\em Sign Languages: Spinning and Unraveling the Past, Present and
  Future. 9th Theoretical Issues in Sign Language Research Conference,
  Florianopolis, Brazil}, volume 382.

\bibitem[\protect\citename{Ong and Ranganath}2005]{ong_automatic_2005}
{Sylvie C. W.} Ong and Surendra Ranganath.
\newblock 2005.
\newblock Automatic sign language analysis: a survey and the future beyond
  lexical meaning.
\newblock {\em {IEEE} Transactions on Pattern Analysis and Machine
  Intelligence}, 27(6):873 -- 891, June.

\bibitem[\protect\citename{Stokoe}2005]{stokoe_sign_2005}
William~C. Stokoe.
\newblock 2005.
\newblock Sign language structure: An outline of the visual communication
  systems of the american deaf.
\newblock {\em Journal of Deaf Studies and Deaf Education}, 10(1):3--37,
  January.

\bibitem[\protect\citename{Valli and Lucas}2000]{valli_linguistics_2000}
Clayton Valli and Ceil Lucas.
\newblock 2000.
\newblock {\em Linguistics of American Sign Language Text, 3rd Edition: An
  Introduction}.
\newblock Gallaudet University Press.

\bibitem[\protect\citename{Wittmann}1991]{wittmann_classification_1991}
Henri Wittmann.
\newblock 1991.
\newblock Classification linguistique des langues signées non vocalement.
\newblock {\em Revue québécoise de linguistique théorique et appliquée},
  10(1):88.

\end{thebibliography}
\end{document}